\documentclass{article}
\usepackage[utf8]{inputenc}
\usepackage{amsmath}
\usepackage{kky}
\usepackage{hyperref}
\usepackage[paper=a4paper,margin=1in]{geometry}% http://ctan.org/pkg/geometry

\usepackage{cleveref}

\newcommand{\minus}{\scalebox{0.75}[1.0]{$-$}}

\usepackage{algorithm}
\usepackage{algorithmic}
\usepackage{subcaption}

\crefname{theorem}{theorem}{theorem}
\crefname{corollary}{corollary}{corollary}
\crefname{assumption}{assumption}{assumption}
\crefname{lemma}{lemma}{lemma}
\crefname{remark}{remark}{remark}
\crefname{proposition}{proposition}{proposition}
\crefname{conjecture}{conjecture}{conjecture}
\crefname{definition}{definition}{definition}
\newtheorem*{theorem*}{Theorem}

\title{Federated Learning as a Mean-Field Game}
\author{Arash Mehrjou\\\href{mailto:amehrjou@inf.ethz.ch}{\small amehrjou@inf.ethz.ch}}
\date{\small ETH Z\"urich \\ \& \\Max Planck Institute for Intelligent Systems}

\begin{document}

\maketitle

\begin{abstract}
    We establish a connection between federated learning, a concept from machine learning, and mean-field games, a concept from game theory and control theory. In this analogy, the local federated learners are considered as the players and the aggregation of the gradients in a central server is the mean-field effect. We present federated learning as a differential game and discuss the properties of the equilibrium of this game. We hope this novel view to federated learning brings together researchers from these two distinct areas to work on fundamental problems of large-scale distributed and privacy-preserving learning algorithms.
\end{abstract}

\section{Introduction}
\label{sec:intro}

In the following sections, we briefly review the necessary material from federated learning and mean-field games that will be used later when similarities start to appear. In federated learning (\Cref{sec:fed_learning}), we present the general idea and one of its standard algorithm called Federated Averaging (FedAvg). The general setting of stochastic games is introduced in~\Cref{sec:stochastic_games}. Two main pillars of mean-field games that are stochastic optimal control and stochastic differential games are introduced in~\Cref{sec:stochastic_optimal_control,sec:stochastic_differential_games} respectively. The introduction of mean-field games is completed in~\Cref{sec:mfg}. Finally, the connection between federated learning and mean-field games is established in~\Cref{sec:fedlearning_mfg}.

\subsection{Federated Learning}
\label{sec:fed_learning}
The concept of federated learning refers to a learning scenario where not all data is available to a single learner, instead is distributed among many devices that can be location-wise distant. Each device together with the fraction of data available to it is called a~\emph{client}. We call the set of all clients participating in this distributed learning process, a~\emph{federation}. Each client learns a local model using its local dataset and contributes to a centralized model called the~\emph{server} at each~\emph{round} of learning. The centralized model is then used to update the client models and the loop continues for the subsequent rounds until some measure of performance is met. 

We are particularly interested in the setting where a large number of clients are participating in the federation. For example, there are hundreds of millions of active mobile phones in the world each of which has its own local dataset from a particular modality (textual, auditory, visual, etc) generated by its user. One of the first commercial applications of federated learning was to use the enormous wealth of textual data generated by users of smartphones to develop a language model to predict the next words while typing. The amount of data generated by each user is not sufficient to develop a satisfactory local model. Hence, it would be useful to use the data provided by millions of other users to learn a better model while preserving the users' privacy by not sending their data to a distant server.

The key constraint in federated learning is that the clients can only communicate via updating a centralized model and then getting updated by it. This model of communication is enforced due to privacy constraints or technical limitations such as limited bandwidth. In the mobile phone example, the data generated by each user, most likely, comes from a distinct distribution depending on the age, location, education, and other attributes of the user. Moreover, the communication is costly as it is normally carried out via the cellphones' internet connection.

Morally speaking, the objective of federated learning is to use locally trained models to find a single centralized model that is as good as the model that could have been trained if all datasets were available in one place to a single learner. Here, we formalize this objective mathematically as it will be needed in the coming sections.

The concept of federated learning can be studied separately from the learning algorithm used by the clients. For convenience, we present the main ideas for a multi-class classification problem where a function is sought to map members of the input space $\Xcal$ to the output space $\Ycal$ consisting of a finite number of values, e.g., $|\Ycal|$=c. Let $\Delta_\Ycal$ be the simplex over $\Ycal$ i.e., $\sum_i o_i=1$ for $o=[o_1, o_2, \ldots, o_k]$. The learning algorithm searches in the hypothesis family $\Hcal$ consisting of hypothesis functions $h\in\Hcal:\Xcal\to\Delta_{\Ycal}$ based on a specified loss function where $h(x)$ is the probability distribution over $[1,2,\ldots, c]$ determining the probability that the input instance $x$ belongs to every class of $\Ycal$. For every hypothesis $h$ and the labeled sample $(x, y)$, the loss function $\ell:\Hcal\times\Delta_{\Ycal}\times \Ycal\to\RR_0^+$ gives a non-negative values $\ell(h(x), y)$. Every \emph{task} is a distribution on $\Xcal\times \Ycal$ denoted by $\Dcal$. The~\emph{risk} (expected loss) of the hypothesis $h$ on task $\Dcal$ is denoted by $\Lcal_{\Dcal}(h)$ where

\begin{equation}
    \label{eq:risk_minimization}
    \Lcal_{\Dcal}(h)=\mathop{\Eb}_{(x, y)\sim\Dcal}[\ell(h(x), y)]
\end{equation}
and $h_{\Dcal}:=\argmin_{h\in\Hcal} \Lcal_{\Dcal}(h)$ is the hypothesis with minimum risk.

Consider a federation of $p$ learners (known as clients in FL literature) indexed by $k\in\{1,2, \ldots, p\}$. The $k$-th learner has the data distribution $\Dcal_k$ where $m_k$ samples of it $S_k=\{(x_{i, k},y_{i, k}),\ldots, (x_{m_k, k},y_{m_k, k})\}$ is available for learning Hence, total number of samples available to all learners is $m=m_1+\ldots+m_p$. Let the hatted notation $\hat{\Dcal}_k$ be the uniform distribution on the samples $S_k$ available to client $k$, known as the empirical distribution. The objective of federated learning is to learn from the $p$ sample sets $\{S_1, \ldots, S_p\}$ a hypothesis $h$ that performs well on some \emph{target} distribution denoted by $\Dcal^t$. The problem falls into different categories depending on the chosen target distribution.

The target distribution is often constructed by a mixture of the client's distributions as
\begin{equation}
    \label{eq:target_distribution_Canonical_fed_learning}
    \bar{\Ucal}=\sum_{k=1}^p \frac{m_k}{m}\Dcal_k,
\end{equation}
where the weight of each client is proportional to its share of the total samples. As a result, the empirical target distribution becomes $\hat{\Ucal}=\sum_{k=1}^p\frac{m_k}{m}S_k$ based on which the actual evaluation is carried out. The server sends the central model to the clients and they use their local data to update their copy of the model. The updated models are then sent to the server and aggregated to build a new central model. This two-phase process repeats until a stopping criterion is met (see~\Cref{alg:canonical_fed_learning}). 

Based on the way client's updates are aggregated, two branches can be thought of. In the first branch, each client produces its update as the gradient $g^k=m_k^{\minus 1}\sum_{x\in S_k}\nabla_w\ell(h(x;w), y)$ computed on its local data. In a non-stochastic setting,  every client uses its whole local data (deterministic gradient descent) to compute its local gradients. If the learning rate is shared across all clients, the central model is updated by averaging the local gradients as $w_{r+1}\leftarrow w_r - \eta \sum_{k=1}^n \frac{m_k}{m}g^k$ . The update strategy of the well-known method $\texttt{FedSGD}$\cite{mcmahan2017communication} is in principle similar to this gradient aggregation rule.

An alternative branch is to apply the gradients locally and then update the central model by averaging over all clients' models. This aggregation method gives more flexibility as the clients can train their local models for multiple epochs and with different learning rates. The local models are updated by $w^k_{r+1}\leftarrow w^k_r - \eta_k m_k^{\minus 1}\sum_{x\in\Bcal} \nabla_w \ell(h(x);w)$ for as many epochs as needed. Once all clients update their models in parallel, the central model is updated by $w_{r+1}\leftarrow \sum_{k=1}^p \frac{m_k}{m} w_r^k$. This method is called \texttt{FedAvg}~\cite{mcmahan2017communication} and its pseudo-code is provided in~\Cref{alg:canonical_fed_learning}. Theoretically speaking, the performance of the output model of \texttt{FedAvg} can be poor when the data distributions of the clients are too different. Intuitively, different distributions can be seen as different tasks and averaging over the weights of the models that solve different tasks can result in a model whose performance is worse in either of them. This effect has been shown by averaging the weights of two digit-recognition networks that are trained on distinct subsets of MNIST long enough that the models begin to overfit to their local data sets~\cite{goodfellow2014qualitatively}.

Both above-mentioned learning methods weigh the client's contribution to the updated central model by $m_k/m$. This is a reasonable weighting choice because, as suggested by the target task in~\Cref{eq:target_distribution_Canonical_fed_learning}, the learned model has to perform well on the mixture of clients' distributions where each client's share in the evaluation dataset is proportional to the size of its local dataset.

\def\NoNumber#1{{\def\alglinenumber##1{}\State #1}\addtocounter{ALG@line}{-1}}
\begin{algorithm*}[!ht]
    \caption{Federated learning by model averaging (FedAvg)}%, after \citep{goodfellow2014generative}.}\label{alg:kgtgan:ours}
    \label{alg:canonical_fed_learning}
    \hspace*{\algorithmicindent} \textbf{Input:} $\{S_1, \ldots, S_p\}: $ Local datasets, $C\in[0, 1]:$ Fraction of clients chosen for each FL update round, $E:$ The number of epochs for training each local model, $\mathtt{batchsize}:$ The size of the minibatches, $\eta:$ learning rate, $\ell:$ loss function, $\mathcal{H}=\{h(\cdot;w)\}$: The hypothesis family, parameterised vector $w$.\\

    \textbf{Server side:}
    \begin{algorithmic}[1]
        \STATE $r (\text{round index}) \leftarrow 0$
        \WHILE{Stopping criterion not met}
        \STATE $m\leftarrow \min(1, Cp)$
        \STATE $M_r\leftarrow$ A random set of $n$ clients chosen at round $r$
        \FOR{each client $k\in M_r$ \textbf{in parallel}}
            \STATE $w^k_{r+1}\leftarrow$ \textbf{ClientUpdate}$(k, w_r)$
            \STATE $w_{r+1}\leftarrow 1/m \sum_{k=1}^n w^k_r$

        \ENDFOR
        \ENDWHILE

    \end{algorithmic}

    \texttt{\\}
    \textbf{ClientUpdate}$(k, w):$
    \begin{algorithmic}[1]
        \STATE $B \leftarrow$ Break $S_k$ into batches of size $\texttt{batchsize}$
        \FOR{each epoch $i:1$ to $E$}
        \FOR{each batch $b\in B$}
        \STATE $w\leftarrow w-\frac{\eta}{|b|}\sum_{(x, y)\in b} \nabla\ell(h(x;w), y)$
        \ENDFOR
        \ENDFOR
    \STATE return $w$
    \end{algorithmic}
\end{algorithm*}

\section{Stochastic Games}
\label{sec:stochastic_games}
The mean-field theory introduced by Lasry-Lions~\cite{lasry2007mean} has attracted a great deal of attention in recent years. The framework is concerned with a system of a large number of rational agents that try to achieve their own goals in the presence of the effect of other agents. In mean-field games (MFG), the agents are all assumed similar in the sense that one agent can be taken as a representative agent of the entire population. 

The difference between MFG and standard optimal control problem is that both the evolution of the states and the decision making of the representative agent is influenced by the whole community of other agents. Inspired by field theory in statistical physics, the effect of the community of agents can be modeled by a mean-field term. Hence the mean-field game boils down to an optimal control problem and a stochastic evolution problem. As a result, the formulation of an MFG can be described as a coupled system of partial differential equations (PDEs). One PDE is the Hamilton-Jacobi-Bellman (HJB) equation that models the evolution for the optimal control part and the other PDE is the Fokker-Plank (FP) equation for the stochastic evolution problem. In this section, we provide a formal mathematical presentation of an MFG setting that will be needed in the coming sections.

A mean-field game is a setting where a large number of agents are interacting each of which tries to achieve its own goal while been influenced by an aggregated effect of the other agents. An agent, as a building block of this population, solves a stochastic optimal control problem. Hence, we first review the setting of stochastic control when the system is governed by It\^{o} dynamics:

\subsection{Stochastic Optimal Control}
\label{sec:stochastic_optimal_control}
We start with assuming the probability space $(\Omega, \Fcal, P)$ where $\Omega$ is the sample space, $\Fcal$ is a sigma field and $P$ is the probability measure. Let $\Fcal_{t\geq 0}$ be a right-continuous filtration and assume an $m-$dimensional $(\Fcal_t)_{t\geq 0}-$adapted Brownian motion is given.

For a measurable space $(U, \Ucal)$, called~\emph{space of actions}, we define the set of~\emph{control strategies} $\UU_t$ for every $t\in[0, T]$ as
\begin{equation}
    \UU_t := \{u:[t, T]\times \Omega\to \Ucal: u\text{ is } \Fcal\text{-optional}\}.
\end{equation}
The states of the agent under the control $u\in\UU_s$ when started from $X_0=x\in\RR^d$ is described by the  It\^{o}'s integral as
\begin{equation}
    X_t = x + \int_s^t b_r(X_r, u_r)dr + \int_s^t\sigma_r(X_r)dW_r
\end{equation}
where $b:[0, T]\times \RR^d\times U\to\RR^d$ and $\sigma:[0, T]\times \RR^d\to \RR^{d\times m}$ are measurable maps. In this expression, $b$ and $\sigma$ are called the~\emph{drift} and~\emph{diffusion} coefficients respectively. We assume the required regularity conditions (See theorem 2.5 of Chapter 5 of~\cite{karatzas2014brownian}) for the existence of the strong solution $(X_t^{s, x, u})_{t\geq 0}$.

We define the instantaneous objective that the agent seeks to maximize, called~\emph{running reward} (or running cost if the agent minimizes it) function as

\begin{equation}
    f:[0, T]\times \RR^d \times U \to \RR
\end{equation}
and a terminal reward that materializes at the final time $T$ as

\begin{equation}
    g:\RR^d \to \RR.
\end{equation}

Both $f$ and $g$ are assumed to be measurable and bounded to avoid pathological situations.

When the agent follows the control strategy $u\in\UU_t$, the~\emph{expected payoff} for a certain $t\in[0, T]$ is defined as

\begin{equation}
    J_t(x, u):= \Eb\left[ \int_t^T f_s(X_s^{t, x, u}, u_s)ds + g(X_T^{t, x, u})\right].
\end{equation}

The value function is defined as the maximal obtainable expected payoff by the control strategies available in $\UU_t$ as
\begin{equation}
    \label{eq:value_function}
    V_t(x) := \sup_{u\in\UU_t}J_t(x, u).
\end{equation}

This definition of the value function makes it a solution of an equation called Hamilton-Jacobi-Bellman (HJB) equation. This result is formalized in the so-called~\emph{verification theorem}. Before presenting the theorem, we define the function $H:[0, T]\times \RR^d\times \RR^d, \times \RR^{d\times d}$, called~\emph{Hamiltonian} as
\begin{equation}
    \label{eq:hamiltonian_stochastic_optimal_control}
    H_t(x, z, \gamma) := \sup_{u\in U}\{f_t(x, u) + \langle z, b_t(x, a)\rangle + \frac{1}{2} \textup{Tr}[\sigma_t\sigma_t\tran(x)\gamma]\}
\end{equation}
where $\langle \cdot, \cdot\rangle$ represents the inner product. Now comes the key theorem of this section that characterises the value function.

\begin{theorem}[Verification Theorem, Thm. 3.5.2 of~\cite{pham2009continuous})]
    \label[theorem]{thm:verification}
    Let $v\in C^{1,2}([0,T]\times \RR^d)$ and its growth is upper bounded by a quadratic function, i.e.
    \begin{equation}
        |v_t(x)|\leq C(1+\lvert x \rvert^2), \;\forall (t, x)\in [0, T]\times \RR^d.
    \end{equation}
    Suppose that $v_T(x)=g(x)$ on $\RR^d$ and $v$ is a solution of the HJB equation
    \begin{equation}
        \label{eq:hjb_equation}
        \partial_t v_t(x) = H_t(x, \partial_x v_t(x), D_{xx}v_t(x)).
    \end{equation}
Then $v=V$, i.e., the function $v$ coincides with the value function as defined by~\labelcref{eq:value_function}.
\end{theorem}

The control can be easily derived from the value function using the following result.

\begin{lemma}[Optimal Control Lemma]
    \label{lem:optimal_control}
    Suppose the conditions of the verification theorem are satisfied and there exists a measurable function $\tilde{u}:[0, T]\times \RR^d\to U$ such that,
    \begin{equation}
        H_t(x, \partial_x v_t(x), D_{xx}v_t(x)) = f_t(x, \tilde{u}_t(x))+\partial_x v_t(x) + \frac{1}{2} \textup{Tr}[\sigma \sigma\tran D_{xx} v_t](x),
    \end{equation}
    then $\tilde{u}$ is an optimal (Markovian) control.
\end{lemma}

In short,~\Cref{thm:verification} states that a sufficiently smooth and growth-bounded solution of the HJB equation~\labelcref{eq:hjb_equation} is the value function of the optimal control problem~\labelcref{eq:value_function} and the optimal control is found by maximizing the Hamiltonian point-wise.

\newcommand{\hspacesubfigs}{1ex}
\begin{figure}[t!]
    \centering
    \begin{subfigure}[t]{0.30\textwidth}
        \centering
        \includegraphics[width=1\textwidth]{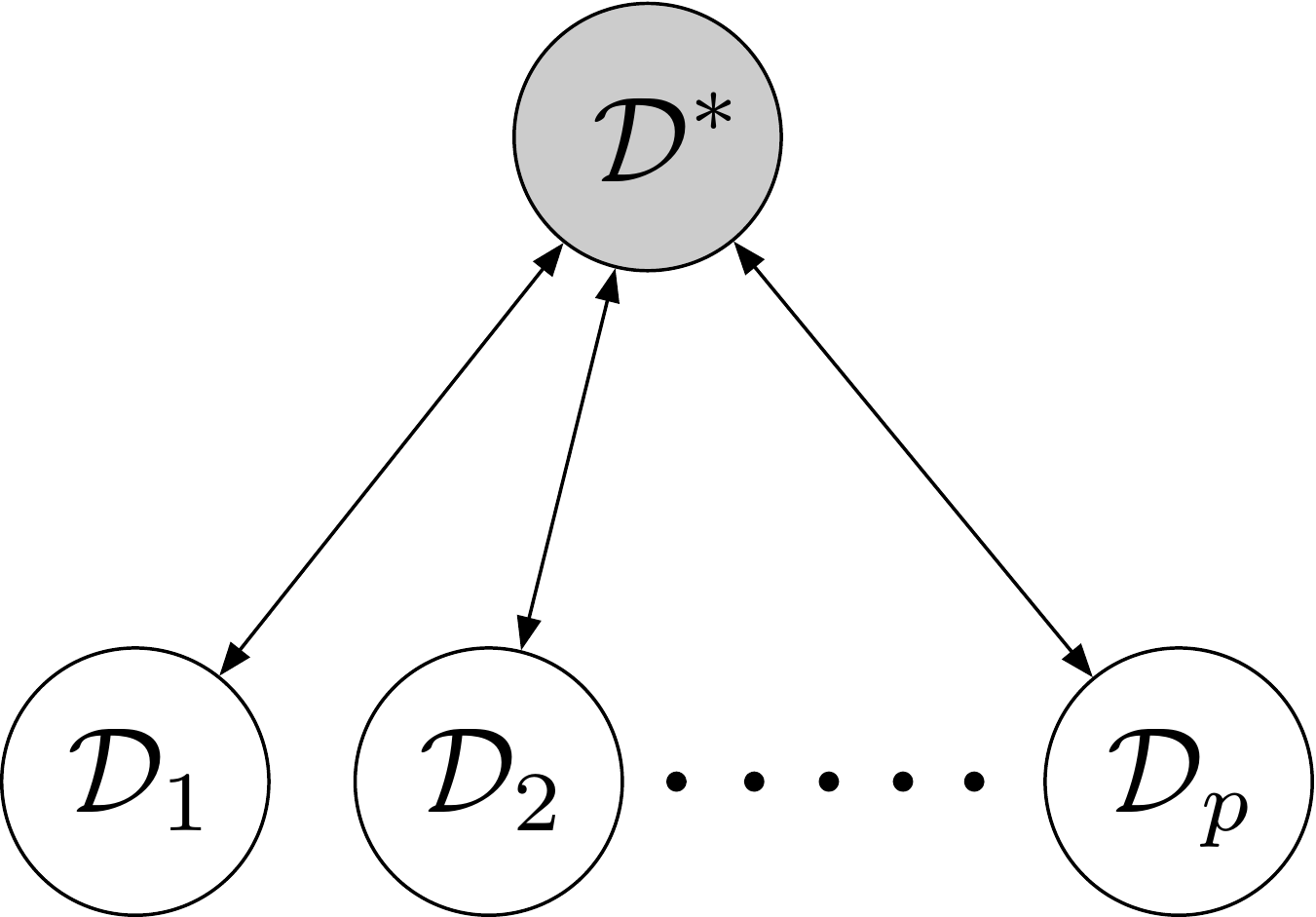}
        \caption{Finite number of clients.}\label{fig:finite_clients}        
    \end{subfigure}
    \hspace{\hspacesubfigs}
    \begin{subfigure}[t]{0.30\textwidth}
        \centering
        \includegraphics[width=1\textwidth]{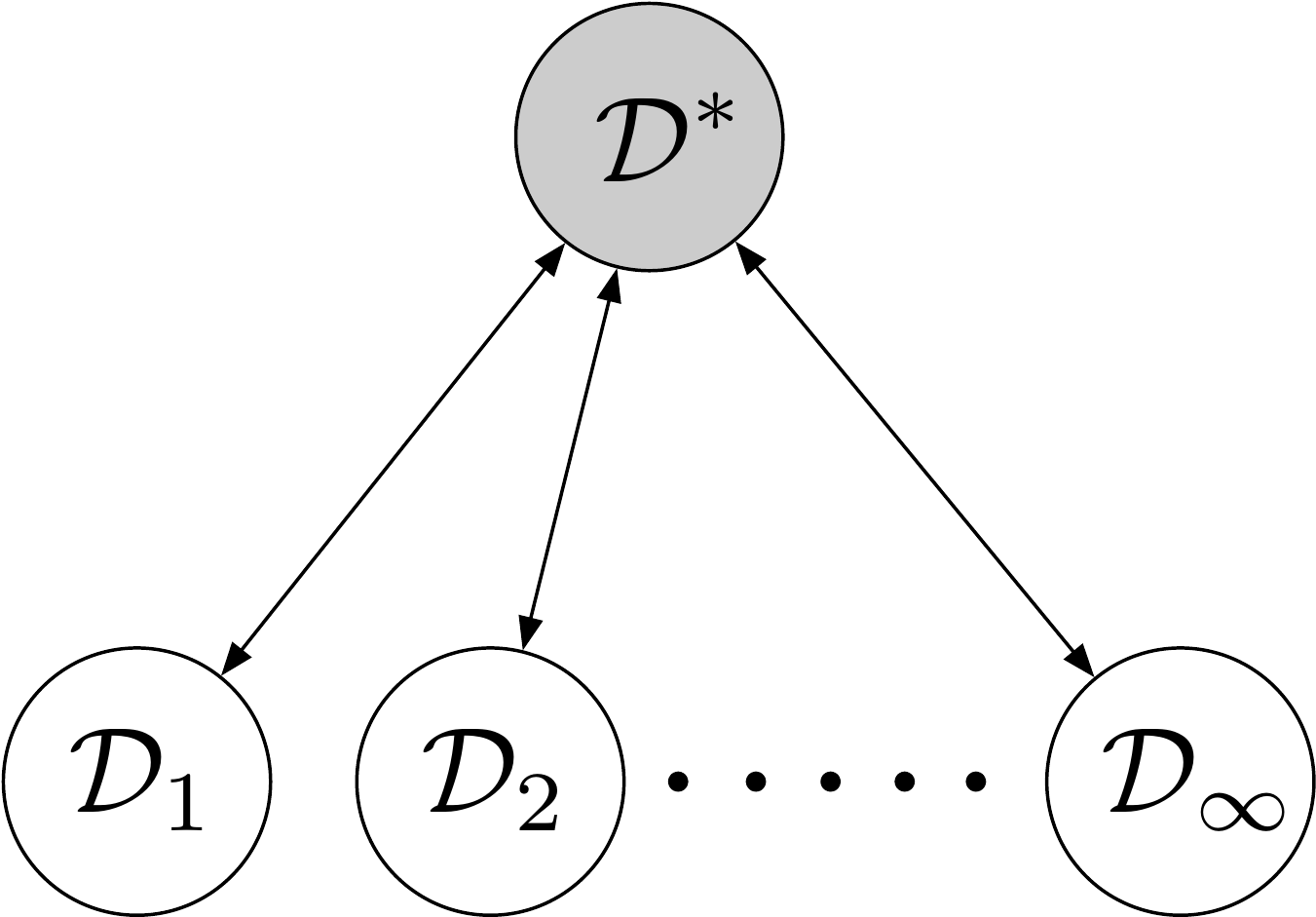}
        \caption{Infinite but countable number of clients.}\label{fig:infinite_countable_clients}        
    \end{subfigure}
    \hspace{\hspacesubfigs}
    \begin{subfigure}[t]{0.30\textwidth}
        \centering
        \includegraphics[width=1\textwidth]{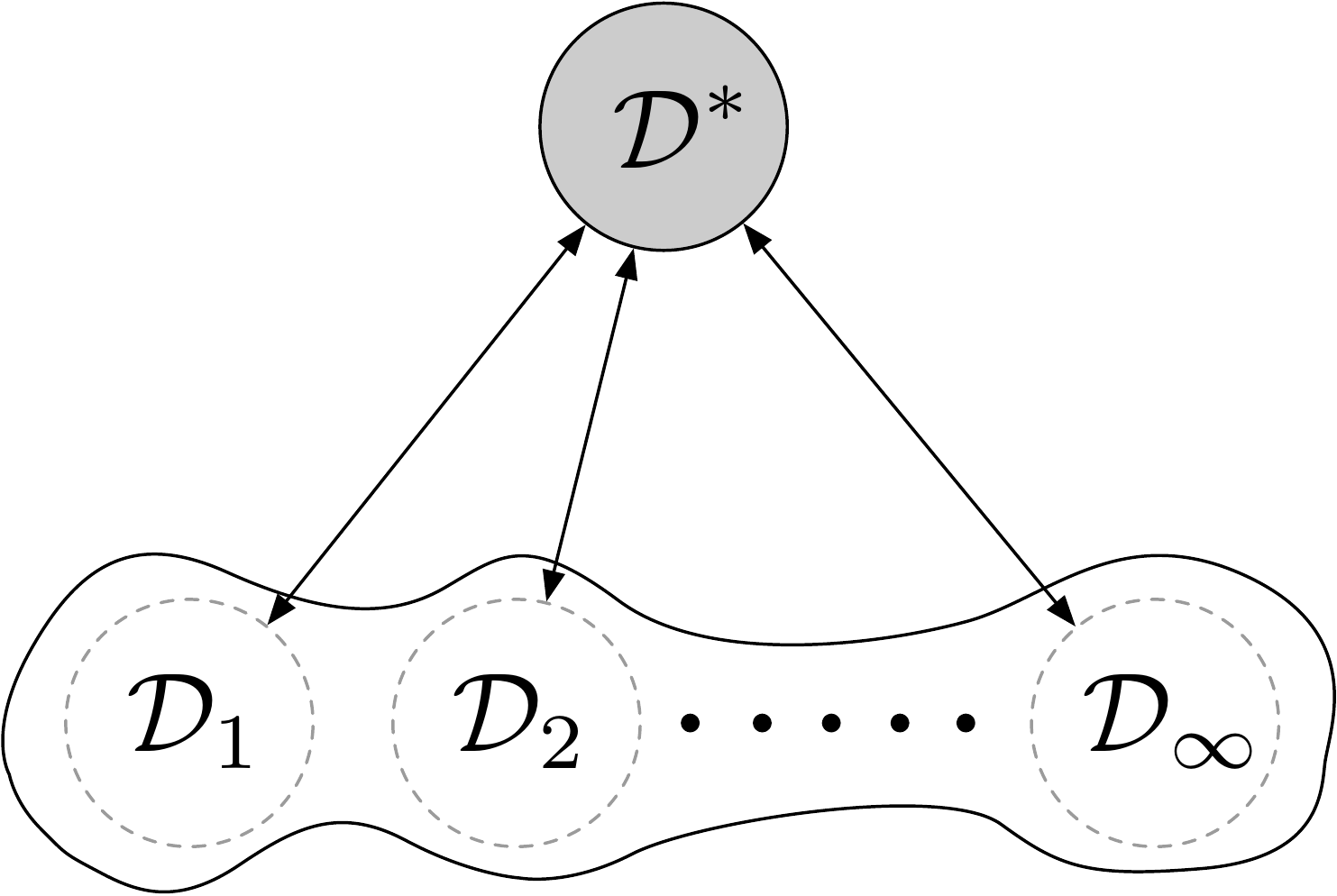}
        \caption{A continuum of clients.}\label{fig:continuum_clients}        
    \end{subfigure}
    \caption{Schematics of federated learning with different cardinality of the set of clients.}

\end{figure}

When we construct the connection between the federated learning and MFG in~\Cref{sec:fedlearning_mfg}, each local client is perceived as an agent who tries to solve an optimal control problem as~\Cref{eq:value_function}. The exact form of the expected payoff $J_t$ and the Hamiltonian $H_t$ for each local learning client will be detailed in~\Cref{sec:fedlearning_mfg}.

To prepare for the presentation of mean-field games in the next section, we first briefly review the concept of stochastic differential games.

\subsection{Stochastic Differential Games}
\label{sec:stochastic_differential_games}
We considered a single agent in~\Cref{sec:stochastic_optimal_control} that controls a stochastic system. The extension of the setting to a finite number of interacting agents is studied in game theory. Assume $p\in\NN$ agents try to maximize their benefits from a common stochastic system. We provide the mathematical formalization of the theory here as a preparation to introduce the mean-field games in the next section.

Let's first fix a few notational conventions. Assume $O$ represents a property such as a state, control, etc. Then, $O_{i,t}$ is the property of the $i$-th agent at time $t$. The notation $O_i$ is the property $O$ for the $i$-th agent for the entire considered time (e.g. $0\leq t\leq T$ or $0\leq t\leq \infty$). A boldfaced notation is used to show the concatenation of the properties for all agents at a certain time. That is, $\Ob_t=(O_{1,t}, \ldots, O_{p, t})$. Hence, the coarsest-grained object is $\Ob$ that encapsulates the property $O$ of all agents at all considered times.

Let $[p]:=\{1, 2, 3,\ldots, p\}$ be the set of $p$ agents (or \emph{players} in this context). Each agent $i\in[p]$ has its own space of strategies 
\begin{equation}
    \UU_{i, t}:=\{u:[t,T]\times \Omega\to U_i : u\text{ is } \Fcal\text{-progressive}\},
\end{equation}
where the pair $(U_i, \Ucal_i)$ is the measurable space of actions for the $i$-th agent and $t\in[0, T]$. By defining the strategies of the agents as stochastic processes, it is important to notice the implicit assumptions on the randomness structure of the game to avoid measure theoretic issues.

\begin{remark} [Information Structure of the game]
    \label[remark]{rmk:information_structure}
    The strategies of all players $\{\UU_1, \UU_2, \ldots, \UU_p\}$ are adapted to the same filteration. That is, there is a mutual source of randomness among all players.
\end{remark}

Let $X_{i,t}\in\RR^d$ represents the $d$-dimensional state of the $i$-th player. By concatenating the states of all $p$ players in $\Xb_t\in\RR^{pd}$, the concatenated state evolution forms a stochastic system that is jointly controlled by all players of the game as
\begin{equation}
    \label{eq:stochastic_differential_game}
    \Xb_t = \xb_s + \int_s^t b_s(\Xb_s, u_{1,s}, u_{2,s}, \ldots, u_{p,s}) + \int_s^t \sigma_s(\Xb_s)dW_s.
\end{equation}

By assuming the necessary requirements on the coefficients $b$ and $\sigma$, there exists a strong solution for~\labelcref{eq:stochastic_differential_game} represented by $(\Xb_t^{s,\xb,\ub})_{t\geq 0}$ that starts from the initial state $\xb_s$ at time $t=s$ where $\ub=(u_1, u_2, \ldots, u_p)$ is the concatenation of the strategies of all players. The payoff for the $i$-th agent at time $t$ is then defined as
\begin{equation}
    \label{eq:payoff_diff_game}
    J_{i, t}(\xb, u_i, u_{-i}) := \Eb\left[ \int_t^T f_{i, s}(\Xb_s^{t,\xb,\ub}, \ub_s)ds + g_i(\Xb_T^{t, \xb, \ub})\right]
\end{equation}
where $f_i$ and $g_i$ are the running and terminal cost for the $i$-th agent. The Nash equilibrium is then defined for this stochastic differential game as the following.

\begin{definition}
\label{def:nash_equilibrium_differetial_games}
    A vector of control strategies $(\tilde{u}_1, \tilde{u}_2, \ldots, \tilde{u}_p)\in \UU_{1,0}\times \UU_{2,0}\times\ldots \UU_{p,0}$ is called a Nash-equilibrium of the stochastic differential game If
    \begin{equation}
        J_{i, o}(\xb, \tilde{u}_i, \tilde{u}_{-i})\geq J_{i, 0}(\xb, u_i, \tilde{u}_{-i})
    \end{equation}
    for every player $i\in[p]$ and every strategy $u_i\in \UU_{i,0}$
\end{definition}

This definition implies that, at a Nash equilibrium, there is no motivation for any agent to change its control strategy while other agents stay with their current strategies. The existence of Nash-equilibrium in a stochastic differential game can be shown by PDE methods (e.g. see~\cite{frehse1984nonlinear}). Dealing with Nash-equilibria becomes quickly intractable when the number of players increases in a strategically interacting game. The mean-field view offers a mathematically delicate way to handle the prohibitively large number of agents.

\subsection{Mean-field games}
\label{sec:mfg}
The interaction among distinct players in a stochastic differential game can quickly become intractable when the number of players increases. A simplifying step is to symmetrize the problem and consider all agents alike. Hence, instead of studying every player separately, one can study a~\emph{representative player} whose interaction with the other players is statistical. As a result, the running cost $f$ and the terminal cost $g$ in the payoff~\labelcref{eq:payoff_diff_game} do not depend on the player index $i\in[p]$. We assume for now that $\sigma$ is an $n\times n$ diagonal matrix. Hence, each player has its own stochastic dynamics whose Brownian motion is independent of the Brownian motion of the other players. Moreover, we made the assumption that a representative player interacts with the other players statistically. This makes the drift $b$ a function of the distribution of players' states instead of each player's action. That is

\begin{equation}
    b: [0, T]\times \RR^d\times \Pcal(\RR^d)\times U\to \RR^d
\end{equation}

where $\Pcal(\RR^d)$ denotes the set of all probability measures defined on $\RR^d$. In the absence of the population distribution of players' states, $b$ is s function of the empirical distribution of states at time $t$ as
\begin{equation}
    \label{eq:empirical_state_distribution}
   \mu_t^p = \sum_{i\in[p]}\delta_{X_{i, t}},\in \Pcal(\RR^d)
\end{equation}

\begin{remark}[Convergence of Random measures]
    \label[remark]{rmk:convergence_of_empirical_probability_measures}
    The empirical distribution $\mu^p$ is a random measure. It is intuitively clear that as $p\to \infty$, this random measure gets closer to a deterministic measure. The conditions under which the random empirical measure converges to a deterministic measure in the narrow topology $P$-a.s. is investigated in Glivenki-Cantelli theorem~\cite{van2000asymptotic}.
\end{remark}

The representative player is modeled by a $d$-dimensional stochastic differential equation on the same stochastic basis $(\Omega, \Fcal, P)$ with a $m$-dimensional Brownian motion. We assume a measure on the space of $\RR^d$-valued continuous functions of time as $\mu\in\Pcal(C([0, T], \RR^d))$. The measure $\mu$ determines how likely every trajectory of the diffusion system is. Let define the time projection map $\pi_t:C([0,T], \RR^d)\to \RR^d$ that takes a sample from the input trajectory at time $t$. This induces a probability measure on the system's states at time $t$ defined as
\begin{equation}
    \mu_t:=\mu\circ \pi_t^{-1}.
\end{equation}

The~\emph{representative player} undergoes a stochastic dynamics as
\begin{equation}
    \label{eq:meanfield_evolution_dynamics}
    X_t=x_s + \int_s^t b_r(X_t,\mu_r,u_r)dr + \int_s^t \sigma_r(X_r)dW_r,
\end{equation}
where $b:[0,T]\times \RR^d\times P(\RR^d)\times U\to\RR^d$ and $\sigma:[0,T]\times \RR^d\to \RR^{d\times m}$.  Similar to~\Cref{eq:stochastic_differential_game}, necessary conditions are assumed for the coefficients $b$ and $\sigma$ so that~\ref{eq:meanfield_evolution_dynamics} has a strong solution. One of these conditions is that $b$ must be Lipschitz with respect to its arguments. To check this condition with respect to the probability measure argument of $b$, the space $\Pcal(\RR^d)$ needs to be endowed with the Wasserstein metric; then the normal Lipschitz condition follows.

If the set of admissible controls for the representative player is shown by $\UU$ defined as
\begin{equation}
    \UU := \{u:[0, T]\times \Omega\to U: u \text{ is } \Fcal\text{-optional}\},
\end{equation}
the expected representative payoff will be defined as
\begin{equation}
    \label{eq:meanfield_payoff}
    J^\mu(x,u):=\Eb\left[\int_0^T f_t(X_t^{0,x,u,\mu}, \mu_t, u_t)dt + g(X_T^{0,x,u,\mu},\mu_T)   \right].
\end{equation}
where $X^{0,x,u,\mu}$ is the unique strong solution of~\labelcref{eq:meanfield_evolution_dynamics}. It is assumed that the running cost $f:[0,T]\times\RR^d\times\Pcal(\RR^d)\times U\to \RR$ and the terminal cost $g:\RR^d\times \Pcal(\RR^d)\to\RR$ are measurable and bounded to avoid pathological cases.

Now, we are ready to define the Nash equilibrium counterpart for the mean-field games.

\begin{definition}
    A mean-field equilibrium is a measure $\mu\in\Pcal(C([0, T], \RR^d))$ on the trajectories $C([0, T], \RR^d)$ such that 
    \begin{itemize}
        \item There exists a strategy (control) $u^*\in\UU$ where
        \begin{equation}
            J^\mu(x, u^*)\geq J^\mu(x,u),\;\forall u\in\UU
        \end{equation}
        \item The probability measure $\mu$ is the law of the state trajectories of the representative player $X^{0, x, u^*,\mu}$ under the strategy $u^*$, that is,
        \begin{equation}
            \label{eq:meanfield_equilibrium_law}
            \mu_t = \textup{Law}(X_t^{0, x, u^*,\mu}).
        \end{equation}
    \end{itemize}
\end{definition}

The existence of the mean-field equilibria has been studied under several assumptions~\cite{cardaliaguet2010notes,carmona2018probabilistic}. At the equilibrium, the distribution~\labelcref{eq:meanfield_equilibrium_law} is plugged into~\labelcref{eq:meanfield_payoff} resulting in McKean-Vlasov type dynamics

\begin{equation}
    \label{eq:master_equation}
    X_t=x_s+\int_s^t b_r(X_r, \text{Law}(X_r), u_r)dr + \int_s^t\sigma_r(X_r)dW_r.
\end{equation}

The solution of this equation is an equilibrium of the mean-field game whose existence requires Lipschitz assumption with respect to the Wasserstein metric on $\Pcal(\RR^d)$.

By taking a test function $\phi\in C^\infty(\RR^d)$ and applying It\^o's formula, the distributional evolution of the states, called the Fokker-Planck equation, is derived as

\begin{equation}
    \label{eq:fokker_planck_meanfield}
    \minus\partial_t\mu_t(x) = \textrm{div}_x(b_t(x, \text{Law}(X_t), u_t)\mu_t(x))+\frac{1}{2}D_{xx}^2(\textrm{Tr}[\sigma_t\sigma\tran_t](x)\mu_t(x)),
\end{equation}
that characterizes a flow in the space of probability measures.

It can be seen from~\Cref{thm:verification} and the definition of the Hamiltonian~\labelcref{eq:hamiltonian_stochastic_optimal_control} that for optimal control $\hat{u}$,
\begin{equation}
    \label{eq:drift_hamiltonian}
    b_t(x,\mu_t,\hat{u}_t) = D_z H_t(x, \partial_t v_x(x), D_{xx}v_t(x))
\end{equation}
where $v_t(x)$ is the value function defined similar to~\labelcref{eq:value_function}. Plugging~\labelcref{eq:drift_hamiltonian} in~\labelcref{eq:fokker_planck_meanfield} gives together with the HJB equation~\labelcref{eq:hjb_equation}, two coupled PDEs of the mean-field game.

\begin{remark}
    Heuristically speaking, according to~\Cref{eq:stochastic_differential_game}, at the Nash equilibrium, each player needs to know the action of all other agents. However, in the mean-field game approach, the player only needs to have statistical information about the states of the other players.
\end{remark}

\section{Mean-Field Federated Learning}
\label{sec:fedlearning_mfg}
After introducing the concept of federated learning in~\Cref{sec:fed_learning}, stochastic differential games in~\Cref{sec:stochastic_differential_games} and mean-field games in~\Cref{sec:mfg}, we are ready to make a bridge and show how federated learning (FL) can be perceived as a mean-field game. We will use the notations introduced in all previous parts throughout this section.

In a bird's-eye view, in an FL setting, each client of the federation interacts with the other clients only through its contribution to the server model. We first argue that the clients can be considered as players in a stochastic differential game. 

Suppose that the local model $h$ trained by client $k\in[p]$ is represented by the weight vector $w_k\in\RR^d$. For example, $w_k$ can be the weights of a neural network or the coefficients of the basis functions in a kernel-based method. In connection with MFG, the weights $w_k$ are the states of the differential game. The differential nature of the game comes from the fact that gradient-based training can be seen as an ordinary or stochastic differential equation in the space of the model's weights. There are several causes of randomness in training. Stochastic gradient descent, as the workhorse of ML algorithms, inherits the stochasticity of choosing a single sample or a mini-batch randomly at each iteration. In neural networks, the drop-out layers act as another source of randomness during training. Inspired by~\cite{liu2019neural}, we encapsulate all sources of noise in a Brownian motion type of randomness that makes the weight trajectories fluctuate over the course of training. In practice, each client has access to the empirical distribution (samples from the underlying data-generating distribution). This can also be seen as another source of randomness but we assume it is also incorporated in the Brownian motion stochasticity and write loss functions based on the data-generating distribution for convenience. Hence, the loss function for client $k$ is
\begin{equation}
    \label{eq:risk_minimization_client_k}
    \Lcal_{\Dcal_k}(w)=\mathop{\Eb}_{(x, y)\sim\Dcal_k}[\ell(h(x;w), y)]
\end{equation}
where $\Dcal_k$ is the local data distribution of the client. Hence, the training dynamics of SGD is written as

\begin{equation}
    \label{eq:training_dynamics_client_k}
    dw_{k,t} = \nabla_{w_{k,t}} \Lcal_{D_k}(w_{k,t}) dt + \sigma_{k,t}(w_{k,t})dW(t),
\end{equation}
where $\sigma_k:[0,T]\times \RR^d\to \RR^{d\times m}$ is the diffusion coefficient for client $k$ and $W(t)$ is a $m$-dimensional Brownian motion. Notice that, unlike~\labelcref{eq:stochastic_differential_game}, there is no explicit control term in this system. However, there is an implicit controller in optimization algorithms that determines how the estimated gradient at each iteration is used to update the weights. Suppose $g_{k,t}$ represents the gradient computed by client $k$ at time $t$, that is, $g_{k,t}:=\nabla_{w_{k,t}}\Lcal_{D_k}(w_{k,t})$. The way $g_k$ influences the weight trajectory can be controlled in a generic way as

\begin{equation}
    \label{eq:general_controlled_training_dynamics_client_k}
    dw_{k,t} = b_t(g_{k,t}, u_{k,t}) dt + \sigma_{k,t}(w_{k,t})dW(t),
\end{equation}
where $u_{k,t}\in\Ucal$ is the strategy employed by client $k$ that determines the way $g_{k,t}$ is used to update $w_{k,t}$ at time $t$. A special case of $b_t$ is a linear map that modulates the magnitude of gradients along each dimension of the weight vector. Hence, $b(g, u) := u \times g$ where $g$ is a vector, $u$ is a matrix and $\times$ is a matrix product. In particular, $u_{k,t}=\Lambda_{k,t}:=\textrm{diag}(\lambda_{k,t})=\textrm{diag}(\lambda^{(1)}_{k,t}, \lambda^{(2)}_{k,t},\ldots, \lambda^{(d)}_{k,t})$ and

\begin{equation}
    \label{eq:linear_controlled_training_dynamics_client_k}
    dw_{k,t} = \Lambda_{k,t}g_{k,t} dt + \sigma_{k,t}(w_{k,t})dW(t),
\end{equation}
shows the continuous training dynamics when the motion along each weight dimension is controllable. An incentive behind the control structure of~\labelcref{eq:linear_controlled_training_dynamics_client_k} is an observation in practical applications of federated learning that shows the important effect of the number of epochs each client is trained before the aggregation phase starts. It has been observed that too long or too short training of clients are both detrimental to the outcome of federated learning~\cite{mcmahan2017communication}. Hence, $\Lambda_{k,t}$ is a proxy that emulates the length of training for client $k$ at time $t$. More concretely, by upper bounding $\bar{\lambda}_{k,t}=\max(\lambda^{(1)}_{k,t}, \lambda^{(2)}_{k,t},\ldots, \lambda^{(d)}_{k,t})$, one can control the influence of the dataset available to client $k$ to the aggregated model. 

For example, a client may have a large dataset that makes its contribution to the aggregated model, when the server averages over all the clients, non-negligible; but the dataset of that client becomes faulty or unreliable at some time $t_{\text{fault}}$. Decreasing the control coefficient that acts as a learning rate, in this case, can nullify the effect of this faulty dataset in the aggregated model. The client still contributes to the averaged model but the contribution is based on the best previous state of the model as the training over the faulty dataset was skipped by decreasing the learning rate.

The goal of federated learning is to find a model that performs well on a target distribution. Let $\Dcal^*$ denote the target distribution. Usually, as mentioned earlier in~\Cref{eq:target_distribution_Canonical_fed_learning} of~\Cref{sec:fed_learning}, the target distribution is defined as a mixture of the clients' local distributions, that is

\begin{equation}
    \Dcal^* = \sum_{k=1}^p \alpha_k \Dcal_k
\end{equation}
where $\alphab=(\alpha_1, \alpha_2, \ldots, \alpha_p)$ belongs the simplex of $p$ values. Hence, the ultimate goal of federated learning is to minimize

\begin{equation}
    \label{eq:risk_minimization_target_distribution}
    \Lcal_{\Dcal^*}(w_{\text{server}})=\mathop{\Eb}_{(x, y)\sim\Dcal^*}[\ell(h(x;w_\text{server}), y)]
\end{equation}
with respect to the server model's weight vector $w_\text{server}\in\RR^d$. This loss function cannot be evaluated and optimized directly as $D^*$ is not available in one place. Instead, the gradients or updated models calculated by the clients are aggregated to update the weights of the server model. For example, in the aggregation phase at time $t$
\begin{equation}
    \label{eq:server_weights_dynamics}
    dw_{\text{server},t} = \left(\sum_{k=1}^p \alpha_k g_{k,t}\right) dt.
\end{equation}

The server weight is then sent to the clients to update the local models. As the information content of the differential of the server's weights is the aggregation of the differentials of all clients, we add the mean-field term~\labelcref{eq:server_weights_dynamics} to the weight's dynamics of each client that turns~\labelcref{eq:linear_controlled_training_dynamics_client_k} into

\begin{equation}
    \label{eq:client_dynamics_with_aggregation_term}
    dw_{k,t} = \left(\Lambda_{k,t}g_{k,t} + \sum_{k=1}^p \alpha_k g_{k,t}\right) dt + \sigma_{k,t}(w_{k,t})dW(t).
\end{equation}

As a result, \Cref{eq:client_dynamics_with_aggregation_term} models the whole federated learning process in a single continuous-time SDE where the effect of local learning and aggregation are unified in the drift term. In particular, the drift term at the RHS of~\labelcref{eq:client_dynamics_with_aggregation_term} conforms with the observation that in federated learning, the parameters of the model learned by each client is affected by the local client first, and the weights of all other clients after the models are aggregated on the server and the local models are synchronized to the server model. In the continuous-time domain, the order of local updates and aggregated updates vanishes as both occur continuously. The length of local learning though can be controlled by the controller signal $(\Lambda_t)_{t\geq 0}$ that manages the local learning rates.

One of the first motivating applications of federated learning was predicting the stream of words while typing with the cellphone's keyboard~\cite{mcmahan2017communication}. Likewise, image recognition and captioning algorithms can feed on the large bulk of data that is globally available across millions of cell phones. In these setups, the ordinary federated learning with finite number of clients (\Cref{fig:finite_clients}) can be approximated by a differential game with infinite number of clients (\Cref{fig:infinite_countable_clients}) or even a continuum of clients (\Cref{fig:continuum_clients}). This allows us to move from~\Cref{eq:empirical_state_distribution} to~\Cref{eq:fokker_planck_meanfield} and derive the mean-field formulation with a probability density over the weights of the clients (see~\Cref{rmk:convergence_of_empirical_probability_measures}) instead of considering each client's weight vector separately. 

Let $w_t$ be the weight vector of the representative client in a federated setting with $p\to\infty$. Assume the distribution of the gradients $g_{k,t}$ of all clients at time $t$ is $\mu^g_t$ and the diagonal matrix $\Lambda_t\in\RR^{d\times d}$ is the representative control that modulates the gradients along every weight dimension. Hence, we have

\begin{equation}
    \label{eq:representative_weight_ito_dynamics}
    w_t = w_s + \int_s^t b_r(w_r, \mu_r^g, \Lambda_r) dr + \int_s^t \sigma_r(w_r) dW_r
\end{equation}
whose analogy to the dynamical equation (Fokker-Planck equation) of MFG is clear when compared with~\Cref{eq:meanfield_evolution_dynamics}. In the following, we show that the second equation of MFG (HJB equation)  can also be derived in federated learning.

To derive the HJB equation, one needs to identify an optimal control setup in federated learning. There are various ways one can define a cost for the process of federated learning. Due to the fact that clients are normally distant from each other, the cost of communication with the server is a natural definition of cost in a federated setting. This becomes more critical in applications where the clients are cell phones which operate on battery and communicate with the server through a band-limited and costly internet connection. Moreover, the cell phone devices are in action while their local models are being trained. Hence, not only their final model but also their overall performance over the rounds of federated training matters. In our continuous-time dynamics, the communication rounds in federated learning are interlaced with the gradient dynamics that locally learn the weights of the clients. Hence, the number of communications with the server is approximated by the terminal time $T$ in continuous-time dynamics. Let $(w_t^{0, w_0, \Lambda, \mu})_{t\in[0,T]}$ be the solution of~\labelcref{eq:representative_weight_ito_dynamics} with initial weight $w_0$, control strategy $\Lambda=(\Lambda_t)_{t\in[0,T]}$, and the flow of probability measures $\mu=(\mu_t)_{t\in[0,T]}$. We fix $T$ and define the running cost $f_t$ as the risk of the server model on distribution $D^*$ at time $t$, that is

\begin{equation}
    f_t (w) = \Lcal_{\Dcal^*}(w)=\mathop{\Eb}_{(x, y)\sim\Dcal^*}[\ell(h(x;w), y)].
\end{equation}

This definition for the running cost emphasizes our desire that we want the model to perform well not only at the end but also during the course of federated learning. This is especially the case in infinite-horizon learning applications, where learning is always happening due to various reasons such as the non-stationarity of the local datasets. For example, the data distribution on mobile phones may change over time as the users' behavior and habits change.

Hence, the expected payoff becomes

\begin{equation}
\label{eq:generic_transiet_and_asymptotic_cost_of_federated_learning}
    J^\mu(x, \Lambda) := \Eb \left[ \int_0^T f_t(w_t^{0, x, \Lambda, \mu})+ g(w_T^{0, x, \Lambda, \mu},\mu_T)\right],
\end{equation}
where $g$ is the cost of the final model when the federated learning process is concluded. This is a generic formulation that can be specialized to various scenarios. For example, if we are only concerned with the performance of the final trained model, we can simply set $g = f_T$ and $f_t=c$ for $t\in[0, T)$ where the constant $c$ shows the relative importance of learning in the fewest federated update rounds. Having the payoff function, it is straightforward to construct the Hamiltonian and the HJB equation. These steps are the same as~\Cref{eq:hamiltonian_stochastic_optimal_control,eq:drift_hamiltonian}, therefore, we do not repeat them here. 

Inspired by~\Cref{def:nash_equilibrium_differetial_games}, the equilibrium of this game is a set of strategies $\{u_1, u_2, \ldots, u_p\}$ for the local clients where no learner is willing to change its strategy unilaterally. In the special case of~\Cref{eq:client_dynamics_with_aggregation_term} where the control strategy is modifying the learning rate multiplicatively by time-dependent functions $\{(\Lambda_{1,t})_{t\in[0,T]}, (\Lambda_{2,t})_{t\in[0,T]}, \ldots, (\Lambda_{p,t})_{t\in[0,T]}\}$, at the Nash equilibrium, changing the control function $(\Lambda_{k,t})_{t\in[0,T]}$ by the $k$-th local learner increases the cost. Notice that this equilibrium concerns a more general objective than the final model learned by the federated learning process and is consequently defined in a space different from the hypothesis family $\Hcal$. It is defined in $([0, T]\to \Ucal)^k:=([0, T]\to \Ucal)\times ([0, T]\to \Ucal)\times \ldots \times ([0, T]\to \Ucal)$, the space of functions that control the schedule of learning for every local learner such that the transient and asymptotic performance of the aggregated model (output of the federated learning at time $t$ during training) is optimal in the sense of the generic cost defined by~\labelcref{eq:generic_transiet_and_asymptotic_cost_of_federated_learning}.

It was shown in this section that constructing the dynamical system of federated learning as a combination of local learning dynamics and federated aggregation, results in a system of equations similar to those that are derived in mean-field games. This result highlights the bridge we aimed to build between federated learning and mean-field games.

\section{Conclusion}
\label{sec:conclusion}

Federated learning as a distributed and privacy-preserving training method is currently receiving considerable attention due to the omnipresence of the edge devices such as mobile phones and the amount of data they create on daily basis. In addition to the complexity of the theoretical analysis of local learning methods which is the case in any non-distributed deep learning algorithm, the distributed learning and aggregation of the model updates make federated learning twofold complicated for theoretical analysis. To abstract the high-level and dynamical aspect of federated learning from the subtleties of learning the local models, we formalized every local learning process as a dynamical system. The aggregation that is performed at each round of federated learning is modeled as a mean-field effect where each local learner is affected by an average (mean-field) effect of the other learners. Therefore, we showed that the entire process can be modeled as a mean-field game, a topic that is studied in physics, game theory, probability theory, and optimal control. We hope mean-field games opens up a new set of theoretical tools to understand the behavior of distributed learning algorithms, especially, federated learning.

% Hence, the training dynamics for client $k$ is written as
% \begin{equation}
%     \label{eq:risk_minimization}
%     \Lcal_{\Dcal}(h)=\mathop{\EE}_{(x, y)\sim\Dcal}[\ell(h(x), y)]
% \end{equation}

% \clearpage
\bibliographystyle{unsrt}
\bibliography{refs}

\end{document}